\documentclass{article}
\usepackage{spconf,amsmath,graphicx,amsfonts}
\usepackage{multirow, booktabs, makecell}
\usepackage{hyperref}
\usepackage[table,xcdraw,usenames,dvipsnames]{xcolor}
\usepackage[normalem]{ulem}
\usepackage{float}

\newif\ifdraft
\draftfalse

\definecolor{dkgreen}{RGB}{0,179,36}
\definecolor{dkred}{RGB}{240,0,0}
\definecolor{dkblue}{RGB}{0,70,160}
\definecolor{dkorange}{RGB}{230,100,20}
\definecolor{dkbrown}{RGB}{170,50,0}
\definecolor{chestnut}{rgb}{0.8, 0.36, 0.36}
\definecolor{pink}{RGB}{255,0,247}
\definecolor{amber}{rgb}{1.0, 0.75, 0.0}
\definecolor{amethyst}{rgb}{0.6, 0.4, 0.8}

\newcommand{\kl}[1]{\ifdraft \textcolor{BlueViolet}{[{\it KL: #1}]} \fi}
\newcommand{\cmc}[1]{\ifdraft \textcolor{dkbrown}{[{\it CMC: #1}]} \fi}

\ifdraft
\newcommand{\kledit}[1]{\textcolor{RoyalBlue}{{#1}}}
\newcommand{\cmcedit}[1]{\textcolor{olive}{{#1}}}

\newcommand{\klremove}[1]{\textcolor{RoyalBlue}{{\sout{#1}}}}
\newcommand{\cmcremove}[1]{\textcolor{olive}{{\sout{#1}}}}
\newcommand{\mzremove}[1]{\textcolor{amber}{{\sout{#1}}}}
\newcommand{\jcremove}[1]{\textcolor{dkgreen}{{\sout{#1}}}}
\newcommand{\todo}[1]{\textcolor{dkred}{[{\it ToDo: #1}]}}
\newcommand{\note}[1]{\textcolor{dkred}{[{\it Note: #1}]}}
\else
\newcommand{\kledit}[1]{\textcolor{black}{{#1}}}
\newcommand{\cmcedit}[1]{\textcolor{black}{{#1}}}

\newcommand{\klremove}[1]{{}}
\newcommand{\cmcremove}[1]{{}}
\newcommand{\mzremove}[1]{{}}
\newcommand{\jcremove}[1]{{}}
\newcommand{\todo}[1]{}
\newcommand{\note}[1]{}
\fi

\title{Few-Shot Spoken Language Understanding via Joint Speech-Text Models}
\name{
Chung-Ming Chien $^1$ \quad
Mingjiamei Zhang $^2$ \quad 
Ju-Chieh Chou $^1$ \quad
Karen Livescu $^1$
}
\address{
Toyota Technological Institute at Chicago $^1$ \quad
The University of Chicago $^2$ 
}
\begin{document}
\copyrightnotice{979-8-3503-0689-7/23/\$31.00~\copyright2023 IEEE}
\ninept
\maketitle
\begin{abstract}
\vspace{-0.5mm}
Recent work on speech representation models jointly pre-trained with text has demonstrated the potential of improving speech representations by encoding speech and text in a shared space. 
In this paper, we leverage such shared representations to address the persistent challenge of limited data availability in spoken language understanding tasks. 
By employing a pre-trained speech-text model, we find that models fine-tuned on text \cmcremove{data }can be effectively transferred to speech testing data. 
With as little as 1 hour of labeled speech data, our proposed approach achieves comparable performance on spoken language understanding tasks (specifically, sentiment analysis and named entity recognition) when compared to previous methods using speech-only pre-trained models fine-tuned on 10 times more data.
Beyond the proof-of-concept study, we also analyze the latent representations. 
We find that the bottom layers of speech-text models are largely task-agnostic and align speech and text representations into a shared space, while the top layers are more task-specific.
\end{abstract}

\vspace{-2.5mm}
\section{Introduction} 
\vspace{-2mm}

\klremove{\cmcedit{Recent}\cmcremove{In the last few} years, s}\kledit{S}elf-supervised speech representations have emerged as an important tool for improving performance and data-efficiency in various speech applications~\cite{riviere2020unsupervised, hsu2021hubert, baevski2020wav2vec}.
These representations encode linguistic content, prosody variations, speaker characteristics, and semantic information~\cite{mohamed2022self}.
Through \cmcremove{a simple }discretization\cmcremove{ step}, such representations even show text-like properties.
For example, some discretized speech representations are closely related to phonetic units~\cite{hsu2021hubert} while being less sensitive to speaker identity changes~\cite{polyak2021speech}, making them suitable for tasks such as language modeling and speech translation~\cite{lakhotia2021on, lee2022direct, lee2022textless}.  Such representation models also appear to encode word-level and syntax information~\cite{shen2023wave,pasad2023what}.
This combination of properties positions self-supervised speech representations as a bridge between surface-form speech signals and the underlying semantic space.

Building upon these observations, jointly pre-trained speech-text models have been developed with the goal of mapping speech and text into a shared representation space, further facilitating the connection between learned speech representations and written language~\cite{ao2022speecht5, bapna2021slam, zhang2023speechlm, zhang2022speechut}. 
Speech-text joint pre-training has proven helpful for speech recognition, synthesis, and translation\cmcremove{tasks}~\cite{bapna2021slam, zhang2023speechlm, zhang2022speechut,wang2023neural}.  However, our understanding of how these models integrate spoken and written language remains limited, and they have not yet been applied to many spoken language understanding (SLU) tasks.

In this paper, we investigate three speech-text models --- SpeechLM-P, SpeechLM-H~\cite{zhang2023speechlm}, and SpeechUT~\cite{zhang2022speechut}.  We analyze their latent representation space and evaluate their performance on two SLU tasks: speech Sentiment Analysis (SA) and Named Entity Recognition (NER).
Our analysis finds that these models follow a first-align-then-predict pattern~\cite{muller2021first}, similar to the pattern observed in multilingual BERT pre-training~\cite{conneau2019cross}; that is, they encode speech and text in a shared representation space in the first few layers, before making predictions in the remaining layers.
On the SLU tasks, we demonstrate that speech-text models outperform speech-only self-supervised pre-trained models.

We also extend our experiments to few-shot and zero-shot 
settings.
In these settings, we assume limited access to labeled speech data but have more labeled text data, which is generally easier to collect. 
Leveraging the aligned representation space of speech-text models, we fine-tune the models with labeled text data (and limited labeled speech data in the few-shot setting) and evaluate their performance on speech data.
On the SA task, speech-text models exhibit excellent zero-shot cross-modal transferability, matching the performance of models fine-tuned on labeled speech data. 
On the NER task, \cmcedit{there is a larger gap between zero-shot speech-text models and fine-tuned speech models (45.1\% vs. 63.4\% $F_1$ on the SLUE benchmark~\cite{shon2022slue}).}\cmcremove{speech-text models achieve an $F_1$ Score of 48.4\% on the SLUE \cmcremove{NER} benchmark~\cite{shon2022slue} in a zero-shot setting, which is\cmcremove{comes} close}\kl{doesn't seem so close... I might instead say something like "On the NER task, there is a larger gap in the zero-shot setting between speech-text models and fine-tuned speech models (48.4\% vs. 68.1\% F1).
However, with only 1 hour of labeled..."}\cmcremove{to the 68.1\% $F_1$ score of self-supervised speech models fine-tuned on \cmcremove{a much larger amount of}labeled speech data.}
\cmcedit{However, w}\cmcremove{W}ith \cmcremove{as little as }\cmcedit{only} 1 hour of labeled speech data, our proposed approach achieves \kledit{performance} \cmcremove{very }close \klremove{performance compared }to \kledit{that of} previous self-supervised speech models fine-tuned on 10 times more data.

Our main contributions are as follows: \kl{I liked the prevoius phrasing better ("Our main contributions are...") because "include" makes the sentence not grammatical.  If you really prefer "include" then you could say "include the following:"}
(1) 
we show that speech-text models \cmcedit{achieve comparable or better performance than }\cmcremove{outperform }speech-only models on multiple SLU tasks; 
(2) in few-shot and zero-shot settings, we demonstrate speech-text models' transferability from text to speech in \cmcremove{both }SA and NER tasks and achieve close performance to previous work that used full labeled speech data;
(3) we demonstrate the existence of a first-align-then-predict pattern in speech-text models, similarly to multilingual pre-training of text models;
(4) based on the observations above, we design a fine-tuning strategy \cmcremove{, where we freeze}\cmcedit{by freezing} the bottom layers and only updat\klremove{e}\kledit{ing} the top layers, which improves the \cmcedit{zero-shot} performance of speech-text models\cmcremove{ in a zero-shot setting}.
\vspace{-1.5mm}
\section{Related Work}
\vspace{-1.5mm}
\subsection{Few-shot end-to-end spoken language understanding} 
\vspace{-1.5mm}
In this work, we define the task of few-shot end-to-end SLU as \textit{learning an end-to-end SLU model using a small amount of labeled speech data and potentially more labeled text data}.
This research direction has received limited attention so far. 
Previous attempts have involved predicting pseudo-labels for speech data using a pre-trained text-based language understanding model~\cite{pasad2022on, he2023zero}, as well as mapping labeled text data to speech embeddings with a text-to-embedding predictor to generate pseudo speech data~\cite{mdhaffar2022end}.
Another approach to tackle this problem is to combine the supervision signals from multiple SLU tasks to train a multi-task SLU model~\cite{meeus2022multitasl}, which assumes a certain level of similarity between different SLU tasks.

\vspace{-2mm}

\subsection{Self-supervised speech models and speech-text models}
\vspace{-1mm}
Self-supervised speech models are learned from pretext tasks applied to unlabeled speech data, such as masked prediction or contrastive predictive coding~\cite{riviere2020unsupervised, hsu2021hubert, baevski2020wav2vec}. 
Incorporating discretization within the self-supervised learning framework has proven beneficial for downstream tasks, as it \klremove{implicitly aligns}\kledit{seems to align} learned representations with human-defined linguistic units like \cmcremove{words or }phonemes.  \kl{is there evidence of discrete units aligning with words?  If not, then remove it}
For example, HuBERT~\cite{hsu2021hubert} employs k-means clustering to update speech representations iteratively and uses the resulting cluster IDs as training targets.
This approach enables the use of a BERT-style masked prediction loss~\cite{devlin2019bert}, and strengthens the connection between learned speech representations and linguistic units.  This style of pre-trained model has encouraged a series of approaches for learning shared representations between speech and text. 

Recent models that encode speech and text into a shared representation space often rely on supervision signals provided by speech-transcription pairs. 
For instance, SLAM~\cite{bapna2021slam} incorporates a cross-modal masked prediction task defined on speech-transcription pairs.
Similarly, mSLAM~\cite{bapna2022mslam} and MAESTRO~\cite{chen2022maestro} directly use a speech recognition loss.
While SpeechLM~\cite{zhang2023speechlm} and SpeechUT~\cite{zhang2022speechut} do not directly utilize speech-transcription pairs in training, they still rely on a speech-to-token or text-to-token model pre-trained with transcribed speech data. 
Token2vec~\cite{yue2023token2vec} is the only model in this category that does not rely on paired data during pre-training.
However, \klremove{their}\kledit{this} model is not publicly available at the moment, so we cannot further analyze \klremove{this model}\kledit{it} and explore its potential.
In addition to speech-transcription pairs, paired translation data is commonly employed to enhance the models' cross-lingual capabilities~\cite{tang2022unified, cheng2023mu2slam, zhang2023speak, wang2023viola}.
Another line of work explores the effectiveness of speech-text joint training on speech generation models~\cite{wang2023neural, zhang2023speak, wang2023viola, saeki2023virtuoso}.

Speech-text models have shown impressive performance on various speech tasks and potential for scaling up to hundreds of languages~\cite{chen2023maestrou, zhang2023google}.
However, exploration of their capabilities for SLU tasks remains limited.
In addition, while certain models have been found to exhibit cross-modal transfer from speech to text --- for example, a model fine-tuned on speech-to-text translation can be directly used for text-to-text translation~\cite{bapna2022mslam} --- their text-to-speech transfer ability has not been explored. 
\vspace{-3mm}


\subsection{Analysis of cross-lingual self-supervised models}
\vspace{-1.5mm}
\kl{I think that a more natural ordering of the related work would have "Analysis of cross-lingual..." as the last sub-section, since it provides good context but is not as central as the other sub-sections.}
The cross-lingual ability of self-supervised models has been long studied by the natural language processing community~\cite{pires2019how, wu2019beto}.
By pre-training on a multi-lingual corpus, self-supervised text models can learn general knowledge shared across human languages.
Subsequently, when being fine-tuned on a specific language, they can then use the information learned across the pre-training languages and enable zero-shot transfer \cite{wu2019beto}.
Analysis of these multi-lingual models shows that they can be conceptualized as consisting of two main components: a multilingual encoder which aligns different languages into a shared representation space, followed by a task-specific language-agnostic predictor \cite{muller2021first}.
During fine-tuning, the encoder remains almost unchanged, while the predictor learns task-specific knowledge from the supervision signals.
This first-align-then-predict framework provides an explanation for the zero-shot transfer behavior of multi-lingual models.

Relatedly, it has been found that the degree of cross-lingual alignment positively correlates with downstream language-transfer performance \cite{muller2021first}.
In addition, freezing the bottom layers (i.e., the multilingual encoder) during fine-tuning in general only leads to a slight drop in same-language performance \cite{merchant2020what} but potentially improves the cross-lingual ability \cite{wu2019beto}.
\vspace{-3mm}
\section{Methods}
\vspace{-2mm}
\subsection{Pre-trained speech-text models}
\begin{figure}[tbh]
    \centering
    \includegraphics[width=\linewidth]{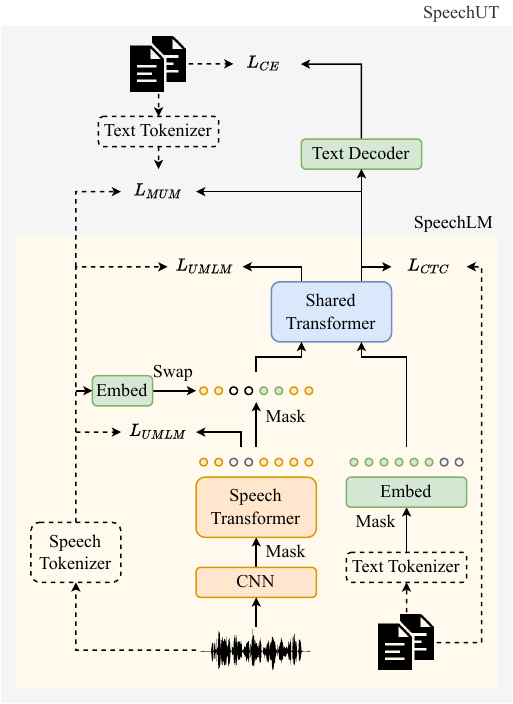}
    \vspace{-2.5mm}
    \caption{The pre-training framework of SpeechLM and SpeechUT. Dashed lines stand for off-line components that are not updated during the pre-training and fine-tuning of the speech-text models.}
    \label{fig:model_arch}
    \vspace{-1.5mm}
\end{figure}

\vspace{-1.5mm}
In this work, we build upon SpeechLM \cmcremove{(SPLM)}~\cite{zhang2023speechlm} and SpeechUT \cmcremove{(\cmcremove{SPUT}\cmcedit{SpeechUT})}~\cite{zhang2022speechut}.\footnote{We use the \textit{Base} configuration for all models.}
As shown in Figure~\ref{fig:model_arch}, the \cmcremove{SPLM}\cmcedit{SpeechLM} model contains two off-line discrete tokenizers for speech and text inputs, a 6-layer speech Transformer and a 6-layer shared Transformer.
The model uses the tokenizers to map speech and text into a shared set of discrete tokens to encourage learning shared representations.
There are two variants of \cmcremove{SPLM}\cmcedit{SpeechLM}, \cmcremove{SPLM}\cmcedit{SpeechLM}-P (P for Phoneme) and \cmcremove{SPLM}\cmcedit{SpeechLM}-H (H for Hidden units), corresponding to different choices of discrete token sets.
\cmcremove{SPLM}\cmcedit{SpeechLM}-P uses phoneme units as the discrete tokens; a speech recognition system and a pre-defined lexicon are used to convert speech signals and text into phoneme units, respectively.
On the other hand, \cmcremove{SPLM}\cmcedit{SpeechLM}-H uses hidden units derived from a k-means model trained on HuBERT~\cite{hsu2021hubert} features as the discrete tokens; the k-means model is also used as the speech tokenizer, while a non-autoregressive text-to-hidden-unit model trained on paired text-unit data is used as the text tokenizer.

Both variants of \cmcremove{SPLM}\cmcedit{SpeechLM} follow exactly the same training procedure, and are trained with a combination of unpaired speech, unpaired text, and a small amount of paired speech-text data.
For speech input, a CNN feature extractor and a 6-layer speech Transformer \cmcremove{is applied and}\cmcedit{are} trained to predict the discrete tokens of masked speech frames in the Unit-based Masked Language Modeling (UMLM) task.
With text token inputs, the shared Transformer is trained with a Unit-based Connectionist Temporal Classification (UCTC) loss to predict the target character sequence.
Like the speech Transformer, the shared Transformer is also trained with the UMLM loss given speech inputs.
To align speech and text representations, a random swapping mechanism is \cmcremove{introduced}\cmcedit{applied} to the inputs of the shared Transformer, where some positions \cmcremove{from}\cmcedit{in} the unmasked region of speech are randomly selected and replaced by tokenized unit embeddings.

The model architecture and training of \cmcremove{SPUT}\cmcedit{SpeechUT} are very similar to \kledit{those of} \cmcremove{SPLM}\cmcedit{SpeechLM}-H as both of them use HuBERT hidden units as the intermediate representations between speech and text.
The main difference is that \cmcremove{SPUT}\cmcedit{SpeechUT} has a text decoder on top of the shared Transformer, which is trained \klremove{on}\kledit{via} Cross Entropy (CE) to predict the target character sequence when given text inputs.
\cmcremove{And t}\cmcedit{T}he shared Transformer is additionally trained on a Masked Unit Modeling (MUM) task to predict masked units given tokenized speech or text as inputs.

All of the models mentioned above are pre-trained on 960~hr\cmcedit{s} of untranscribed speech from the LibriSpeech dataset and 40M text sentences from the LibriSpeech LM corpus~\cite{panayotov2015librispeech}.
The model size, training procedure, and datasets all follow the standard setting of HuBERT-Base~\cite{hsu2021hubert}, which makes our approach directly comparable with previously reported results of HuBERT~\cite{hsu2021hubert} and Wav2Vec 2.0~\cite{baevski2020wav2vec}.
For a fair comparison between \cmcremove{SPUT}\cmcedit{SpeechUT} and other models, we discard the pre-trained decoder in our experiments to ensure a consistent configuration across models.

\vspace{-3mm}

\subsection{Analysis of representation alignment in pre-trained models}
\vspace{-1.5mm}
\label{ssec:analysis_pretrained}
\cmcedit{Knowledge }\cmcremove{The ability to }transfer\cmcremove{ knowledge} from text-based training to speech tasks relies on the shared representation learned during pre-training.
To analyze the alignment between speech and text in the\cmcremove{pre-trained} speech-text models, we use the Average Neuron-Wise Correlation (ANC)~\cite{del2022cross}\kledit{.}
\footnote{Another common model analysis tool is Canonical Correlation Analysis (CCA)~\cite{hotelling1936relations}, which is often applied in model analysis to measure the information shared between two views of the same data instance~\cite{voita2019bottom,pasad2021layer}.
However, CCA allows a linear projection between the two views, and so does not reflect the direct alignment between them.  CCA is therefore more applicable in settings where the information content need not be distributed in the same way across dimensions in the two views.  Empirically, in preliminary experiments in our setting, CCA and ANC analysis largely agree with each other.
}\klremove{.}


Let $X, Y \in \mathbb{R}^{d \times T}$ be \cmcremove{the }two different views\cmcedit{ of the same data instance}, each containing a sequence of $T$ vectors of dimension $d$, representing the activation of $d$ neurons in a given model layer across $T$ time steps.
ANC 
is simply defined as the average correlation of the activations of individual neurons \cmcremove{in the network }$\frac{1}{d}\sum_{i=1}^d corr(X_i, Y_i)$\cmcedit{, with $X_i, Y_i \in \mathbb{R}^{1 \times T}$ being the activation of one single neuron across $T$ time steps}~\cite{del2022cross}. \kl{it is a bit confusing that $T$ doesn't appear in this expression, and it's not stated what $i$ indexes.  can you try to clarify a bit?}

In our experiments, we use paired speech and transcription as inputs to the speech-text models, respectively, and extract the latent representations from each layer of the shared Transformer for the 
ANC computation.
Ground-truth alignments between speech and text sequences are used to ensure frame-wise alignment between the extracted representations.
For \cmcremove{SPLM}\cmcedit{SpeechLM}-P, we employ a pre-trained forced alignment tool~\cite{mcauliffe2017montreal} to obtain accurate phoneme durations and expand the phoneme sequence accordingly given the text inputs.
For \cmcremove{the} \cmcremove{SPLM}\cmcedit{SpeechLM}-H and \cmcremove{SPUT}\cmcedit{SpeechUT}, we apply the HuBERT tokenizer to convert speech signals into discrete tokens and use them as the text inputs.
\cmcedit{With perfectly aligned sequences}\cmcremove{Given sequences perfectly aligned along the time axis}, the 
ANC values then reflect the degree of alignment between the learned text and speech representations, which we view as a prerequisite for the zero-shot and few-shot transferability of speech-text models.

\vspace{-1.5mm}

\subsection{Zero-shot and few-shot spoken language understanding}
\begin{figure}[t]
    \centering
    \includegraphics[width=\linewidth]{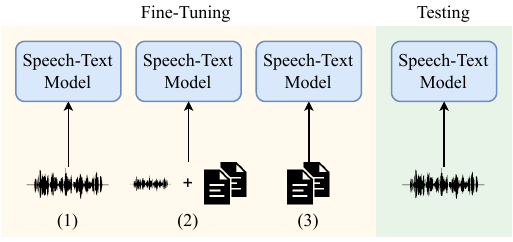}
    \vspace{-2mm}
    \caption{Fine-tuning configurations compared in our work. (1) is the default all-speech fine-tuning setting of the SLUE~\cite{shon2022slue} benchmark, while (2) and (3) refers to the few-shot and zero-shot SLU fine-tuning \cmcedit{studied}\cmcremove{used} in this work. All fine-tuning settings are evaluated with  speech input.}
    \label{fig:diagram}
    \vspace{-1mm}
\end{figure}

\vspace{-1mm}
Figure~\ref{fig:diagram} shows our workflow for fine-tuning speech-text models for spoken language understanding tasks.
In our zero-shot SLU experiments, we fine-tune the model solely using labeled text inputs without any speech data.
Following the pre-training of \cmcremove{SPLM}\cmcedit{SpeechLM}-P~\cite{zhang2023speechlm}, we randomly up-sample the phoneme sequence to match the length distribution of discrete speech tokens and fine-tune the model with upsampled phonemes.
For \cmcremove{SPLM}\cmcedit{SpeechLM}-H and \cmcremove{\cmcremove{SPUT}\cmcedit{SpeechUT}}\cmcedit{SpeechUT}, we use the pre-trained text tokenizer to predict hidden units from text, and fine-tune the model with predicted units.

We also explore the few-shot setting with slightly relaxed data scarcity restrictions.
In this scenario, we assume access to all labeled text data used in the zero-shot setting, as well as a small fraction of the labeled speech data. 
Both types of data are combined for joint fine-tuning of the speech-text models.
Due to the imbalanced data sizes between speech and text, we apply temperature sampling~\cite{arivazhagan2019massively} to increase the likelihood of speech data being sampled during training\cmcremove{ iterations}. 
Specifically, \cmcedit{let}\cmcremove{with} $(p_s, p_t)$ denot\cmcedit{e}\cmcremove{ing} the ratio of speech and text sentences in the dataset, we re-balance the probability of speech and text batches being sampled to $\left(\frac{p_s'}{p_s'+p_t'}, \frac{p_t'}{p_s'+p_t'}\right)$ with $p'=p^{\frac{1}{T}}$.
Following prior work, we set $T=5$ across all few-shot experiments~\cite{arivazhagan2019massively}.

To provide a comparison with our zero-shot and few-shot SLU methods, we establish two performance baselines.
The first baseline involves fine-tuning SpeechLM using all of the available labeled speech data.
While the second baseline model is created by fine-tuning SpeechLM with synthesized speech data. 
The second baseline simulates the zero-shot scenario, where labeled text data is available but no corresponding speech data exists. 
We assume the presence of an additional speech synthesis model \footnote{We use the ESPnet implementation~\cite{inaguma2020espnet} of VITS~\cite{kim2023conditional} pre-trained on the LibriTTS~\cite{zen2019libritts} corpus and randomly sample speakers to generate a synthesized speech dataset with the same size as the original speech dataset.} that can be used to convert the text-label pairs into speech-label pairs.

Inspired by previous layer-wise analyses of self-supervised models which revealed that different aspects of spoken and written languages are encoded in different layers~\cite{muller2021first, pasad2021layer}, we also investigate fine-tuning only the top Transformer layers while keeping the bottom layers frozen~\cite{merchant2020what}. 
This exploration allows us to examine how the performance of the model is affected by fine-tuning specific layers and gain insights into the encoding of speech and text representations in different layers of the Transformer model.

\vspace{-3mm}
\subsection{Analysis of latent representations after fine-tuning}
\vspace{-1.5mm}
In Sec. \ref{ssec:analysis_pretrained}, we \klremove{apply}\kledit{discussed the application of }
ANC to analyze the degree of alignment between speech and text representations in the pre-trained models, as a presumed prerequisite for zero-shot transfer.
Additionally, we \cmcremove{are also interested in understanding }\cmcedit{also want to understand }how the model learns through different fine-tuning setups.
By comparing the ANC between speech and text representations in the fine-tuned model against the pre-trained model, we can examine whether the aligned representation space is preserved after fine-tuning.
On the other hand, we would also like to know how the fine-tuning tasks and input \cmcedit{modalities }\cmcremove{forms}(\klremove{e.g. }speech or text) result in different behavior of speech-text models, and whether speech-text models follow the ``first-align-then-predict" pattern observed in multi-lingual text models~\cite{muller2021first}.
To answer this question, we apply ANC analysis to compare the latent representations of (1) models fine-tuned on the same tasks but with different input \cmcedit{modalities}\cmcremove{forms} and (2) models fine-tuned on different tasks with the same input \cmcedit{modality}\cmcremove{form}, which allow\kledit{s} us to discern the task-agnostic and task-specific components within speech-text models.

\vspace{-2.5mm}
\section{Experiments}
\cmcremove{
}
\vspace{-2.5mm}

\subsection{Layer-wise ANC analysis of pre-trained models}
\begin{figure}[tbh]
    \centering
    \includegraphics[width=\linewidth]{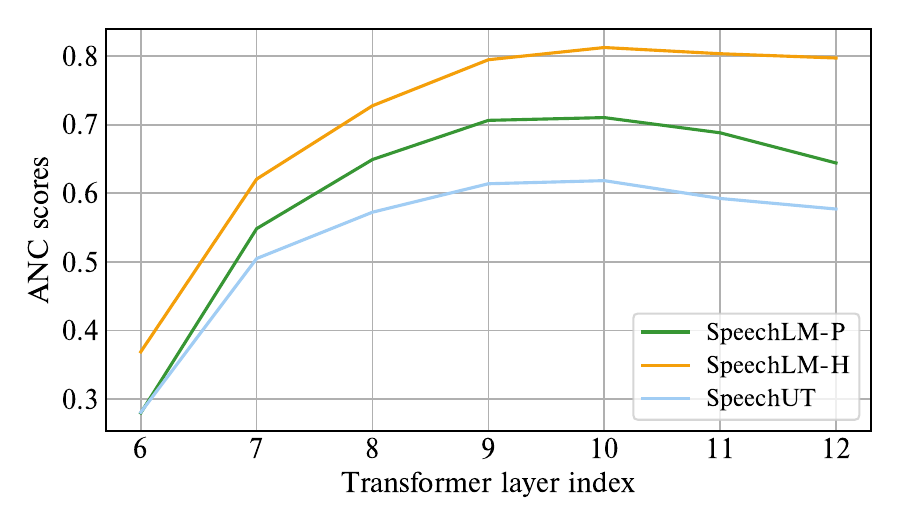}
    \vspace{-4mm}
    \caption{ANC analysis of speech and text representations in three pre-trained speech-text models. ``Layer 6" refers to the input of the shared Transformer.}
    \label{fig:cca_anc}
\end{figure}

\vspace{-2mm}
\label{ssec:anc_pretrained}
To assess the similarity between paired speech and text representations in the shared Transformer of pre-trained speech-text models, we use the dev-clean \klremove{split}\kledit{subset} of the LibriSpeech dataset~\cite{panayotov2015librispeech}.
From this \klremove{split}\kledit{set}, we randomly select 500 utterances, calculate the correlation at each dimension, and then report averaged values.
The results are shown in Fig.~\ref{fig:cca_anc}.

As the the results of the ANC analysis show, despite the inputs of the shared Transformer not being perfectly aligned (in layer 6), the models effectively learn to map text and speech into a shared representation space, as evidenced by the \cmcedit{high} \cmcremove{increasing} correlation scores \cmcremove{across layers. 
Notably, the highest alignment occurs} at around layers 9 and 10. 
However, in the final layers, a decline in 
ANC scores can be observed in all models, which might be attributed to the utilization of distinct pre-training losses for speech and text inputs.
Overall, the observed trend verifies that the bottom few layers in the shared Transformer are capable of aligning speech and text representations in a shared representation space.

\vspace{-2mm}
\subsection{SLUE Sentiment Analysis}
\vspace{-1.5mm}
To assess the cross-modal transferability of the jointly pre-trained speech-text models, we employ the Spoken Language Understanding Evaluation (SLUE) benchmark~\cite{shon2022slue}, which includes \klremove{an SA}\kledit{a sentiment analysis (SA)} task and \klremove{an NER}\kledit{a named entity recognition (NER)} task. 
To fine-tune the \cmcremove{speech-text }models on SA, we follow the standard setup in the SLUE toolkit:
A self-attention pooler is added on top on the pre-trained model to produce a fixed-dimensional feature \kledit{vector} from inputs with variable lengths, followed by a 2-layer classifier trained with cross entropy loss to predict the sentiment class label.
The model is trained for 30k updates with a batch size of 1.4M speech frames.
For text inputs, we set the batch size to 4375 tokens to ensure roughly equal \cmcedit{numbers of sentences in }\cmcremove{sizes }\kl{in terms of number of words?}\cmcremove{ of }\cmcremove{the }speech and text batches.

To evaluate the SA fine-tuning, we report the macro-averaged $F_1$ scores on the \cmcedit{dev subset} \cmcremove{dev split (since the ground-truth labels of the test split are not released) }under our few-shot and zero-shot settings.
A performance evaluation with comparison to prior work is shown in Table~\ref{tab:SA_results}.
When fine-tuned on speech inputs, speech-text models already show better performance than speech-only pre-trained models (44.8--45.6\% vs. 43.3--44.0\%).\kl{what does $\sim$ mean here?  Why not "--"?}
In the zero-shot setting, the models demonstrate excellent transferability from text to speech inputs\cmcedit{ with SpeechUT achieving }\cmcremove{ with }an $F_1$ score of \cmcedit{47.0\%}\cmcremove{45.2$\sim$47.0\%}, which \cmcedit{outperforms all speech-text models and speech-only models fine-tuned on speech.}\cmcremove{not only outperforms prior work (44\%) but also matches the performance of speech-text models fine-tuned on speech (44.8$\sim$45.6\%).}
However, it is worth noting that combining a small amount of speech data with text does not help the models improve the SA performance, which may result from the interference between training signals.

\begin{table}[t]
    \setlength{\tabcolsep}{2pt}
    \centering
    \caption{$F_1$ scores  (\%) of the Sentiment Analysis task on \kledit{the} SLUE dev set.
    \cmcremove{Both baselines and proposed approaches show \kledit{performance} comparable \kledit{to} or better \klremove{performance to}\kledit{than} the prior work\klremove{s}. \kl{table captions should generally go above tables (for figures they go below)}}}
    \footnotesize
    \begin{tabular}{l c r r r r r}
        \toprule
        \multicolumn{2}{c}{Labeled Data
        } & \multicolumn{5}{c}{Models
        } \\
        \cmidrule(lr){1-2}
        \cmidrule(lr){3-7}
        \multirow{2}{*}{Speech} & \multirow{2}{*}{Text} & W2V2 & HuBERT & \cmcremove{SPLM}\cmcedit{Speech-} & \cmcremove{SPLM}\cmcedit{Speech-} & \cmcremove{SPUT}\cmcedit{Speech-} \\
        & & \cite{shon2022slue} & \cite{shon2022slue} & \cmcedit{LM-P} & \cmcedit{LM-H} & \cmcedit{UT} \\
        \midrule
        \textit{Baselines} \\
        \cmcedit{\hspace{2mm} 1 hr} & - & & & 36.9 & 37.7 & 39.6 \\
        \hspace{2mm} 12.8 hrs & - & 43.3 & 43.0 & 45.6 & 45.3 & 44.8 \\ 
        \hspace{2mm} 12.8 hrs (syn) & - & & & 46.4 & 46.3 & 46.1 \\ 
        \midrule
        \textit{Proposed} \\
        \hspace{2mm} - & full & & & 45.2 & 45.2 & 47.0 \\
        \hspace{2mm} 10 mins & full & & & 45.2 & 38.3 & 39.5 \\ 
        \hspace{2mm} 1 hr & full & & & 46.4 & 43.4 & 45.4 \\
        \bottomrule
        \multicolumn{7}{l}{\cmc{We need to reduce fontsize to make space for the fullnames of the models (i.e. SpeechLM \& SpeechUT).}\kl{could increase font size and split "SpeechLM-P" across 2 lines ("Speech-" on the first line), and similarly for SpeechLM-H.}}
    \end{tabular}
    \label{tab:SA_results}
    \vspace{-3.5mm}
\end{table}
\vspace{-2.5mm}
\subsection{SLUE Named Entity Recognition}
\vspace{-1mm}
\label{ssec:ner}

To set up the model for NER fine-tuning, we follow the default configuration of the SLUE toolkit, which adds a linear layer on top of the pre-trained model and trains the models with a character-level Connectionist Temporal Classification (CTC) loss.
The models are trained for 20k steps with a batch size of 3.2M frames for speech inputs and 10k tokens for text inputs.

We evaluate the NER performance on the \cmcedit{dev set}\cmcremove{dev split} of the SLUE dataset and report \cmcremove{a }\cmcedit{the }micro-averaged $F_1$ \cmcremove{score }\cmcedit{and label-$F_1$ scores. 
Label-$F_1$ score considers only the tag predictions and ignores any misspelling and segmentation error\kledit{s} in speech-to-text conversion.}.
There is an option to utilize an offline 4-gram language model (LM) for decoding the model output.
The language model is trained independently on the fine-tune set and generally improves performance.

We evaluate the NER task performance of different fine-tuning schemes\klremove{using} both with and without a language model.
A detailed performance evaluation on the SLUE dev set is shown in Table~\ref{tab:NER_dev_results}.
\klremove{Unlike}\kledit{Compared to} the SA task, which is a simple classification, the NER task is more complicated and involves decoding the model output into a \cmcremove{sequence of characters}\cmcedit{character sequence}.
For \cmcremove{SPLM}\cmcedit{SpeechLM}, text-only training seems to be insufficient for guiding the model to learn about speech labeling, and has significantly worse performance compared to training with speech data.
However, \cmcremove{SPUT}\cmcedit{SpeechUT} demonstrates impressive zero-shot ability, attaining a text-only $F_1$ score of 48.4\% with the aid of LM decoding.
We also observe a significant improvement in the NER performance by incorporating as little as 10 minutes of speech data alongside the text input, which is about 1/100 the size of the full speech dataset.
Without an LM, the 10~mins \kledit{of} speech \klremove{input}\kledit{data} improves \klremove{the text training} performance from 1.2\% to 34.7\% for \cmcremove{SPLM}\cmcedit{SpeechLM}-H, and with 1~hr \kledit{of} speech\klremove{ input}, the performance \klremove{can be further improved}\kledit{further improves} to 52.0\%\cmcremove{, which is only slightly behind the performance of previous work at \cmcedit{49.6\% and 49.8\%}\cmcremove{54.5\% and 55.0\%}}.
\cmcedit{With 3~hrs \kledit{of} speech\klremove{ input}, the performance \klremove{can be updated}\kledit{improves} to 59.6\%, only slightly behind the 64.0\% $F_1$ score achieved by fine-tuning with \kledit{the} full speech data\cmcedit{set}.}
Similar performance improvement can be seen with LM decoding.
The \klremove{text training} performance for \cmcremove{SPLM}\cmcedit{SpeechLM}-P is improved from 9.3\% \kledit{with text-only training} to 50.4\% with 10~mins \kledit{of} speech data, and \klremove{can be}\kledit{is} further improved to 62.5\% \cmcedit{and 69.7\%} with 1~hr \cmcedit{and 3~hrs} of speech data\cmcedit{, respectively}.

\begin{figure}[t!]
    \centering
    \includegraphics[width=\linewidth]{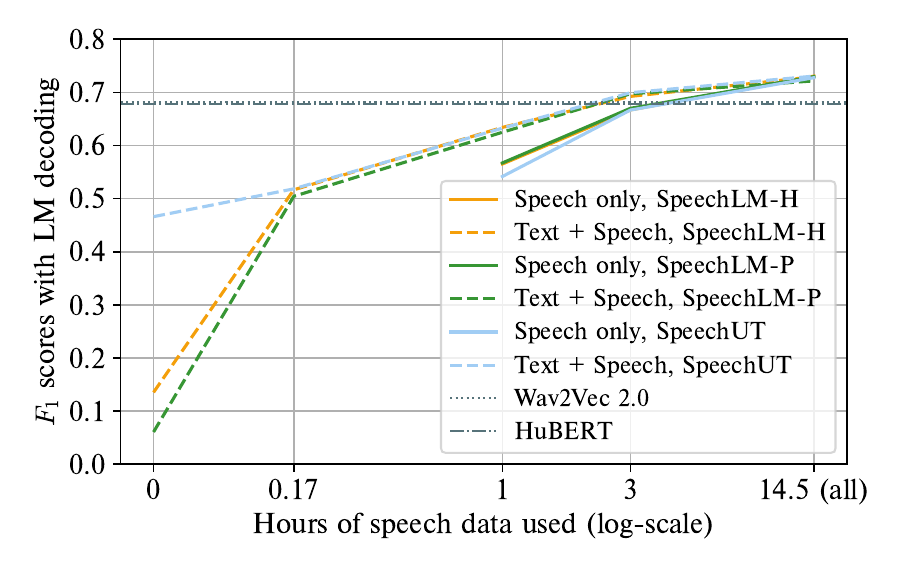}
    \vspace{-5mm}
    \caption{\cmcedit{$F_1$ scores (\%) of the Named Entity Recognition task on SLUE dev set with different amount of speech data used.}}
    \label{fig:NER_hours_speech}
\end{figure}

\cmcedit{
In Fig.~\kl{do you mean Fig?}\ref{fig:NER_hours_speech}, we show the results of fine-tuning speech-text models for the NER task with varying amounts of speech data.
We find that the benefit of using labeled text data is more significant when only limited speech data is available.
With the full labeled speech dataset, text data only results in marginal improvement\klremove{s}.
However, it \kledit{is} worth\klremove{s} noting that the text transcriptions we use to fine-tune the models only correspond to 14.5~hrs of speech data.
The performance can potentially be further improved by using more text data, which is generally easier to collect \klremove{compared to}\kledit{than} labeled speech data. 
}

In Table \ref{tab:NER_test_results}, we show the NER performance of the proposed methods on the SLUE test set 
\footnote{We followed the evaluation protocol to get the test set results on the SLUE benchmark.} with a comparison to prior work.
Similar to the dev set results, SpeechUT demonstrates excellent zero-shot transfer ability with an $F_1$ score of 45.1\% when fine-tuned solely on text.
With 3~hrs of data, we can match the performance of prior work fine-tuned on full speech data (61.9--63.4\%) with any of the speech-text models (62.6--63.8\%).


\vspace{-2.5mm}
\subsection{ANC analysis of fine-tuned models}
\begin{figure}[t!]
    \centering
    \includegraphics[width=\linewidth]{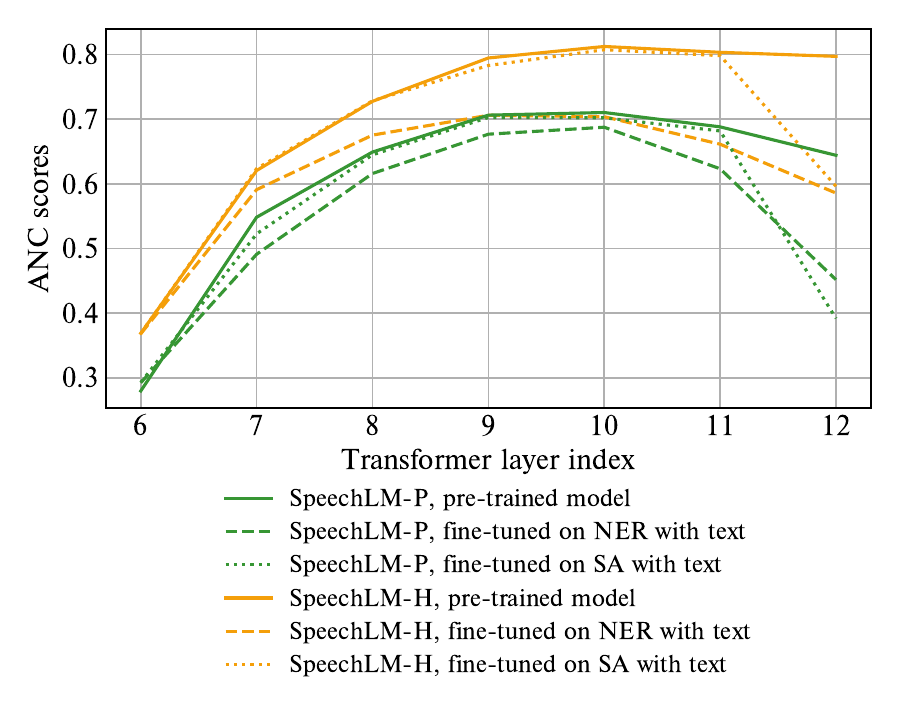}
    \vspace{-5mm}
    \caption{ANC scores between speech and text representations in pre-trained and fine-tuned models. \cmcremove{SPUT}\cmcedit{SpeechUT} results are not shown to avoid \klremove{tangling curves}\kledit{visual clutter}.}
    \label{fig:finetune_speech_text_anc}
\end{figure}

\vspace{-1.5mm}
\label{ssec:anc_finetuned}
We conduct ANC analysis on the latent representations in fine-tuned models to get a clearer picture of how the models learn through fine-tuning.
In Fig.~\ref{fig:finetune_speech_text_anc}, we \klremove{compute}\kledit{show} the ANC \klremove{of}\kledit{between} speech and text representations in pre-trained models and fine-tuned models, respectively.
We follow the same setup as in Sec. \ref{ssec:anc_pretrained} for data preparation.
By comparing the curves of pre-trained \klremove{models against those of}\kledit{and} fine-tuned models, we can see that they almost overlap with each other from layer 6 to layer 10 (with \cmcremove{SPLM}\cmcedit{SpeechLM}-H fine-tuned on NER with text being an exception).
This shows that fine-tuning only marginally affects the speech-text alignment in the latent space of the bottom layers.
After layer 11, pre-training and fine-tuning curves diverge, implying that the top layers are affected more by fine-tuning and thus are more task-specific.
This result, combined with the results in Sec. \ref{ssec:anc_pretrained} and the zero-shot transferability of the models, shows that speech-text models follow the ``first-align-then-predict" pattern observed in multi-lingual text models.

\begin{figure}[t!]
    \centering
    \includegraphics[width=\linewidth]{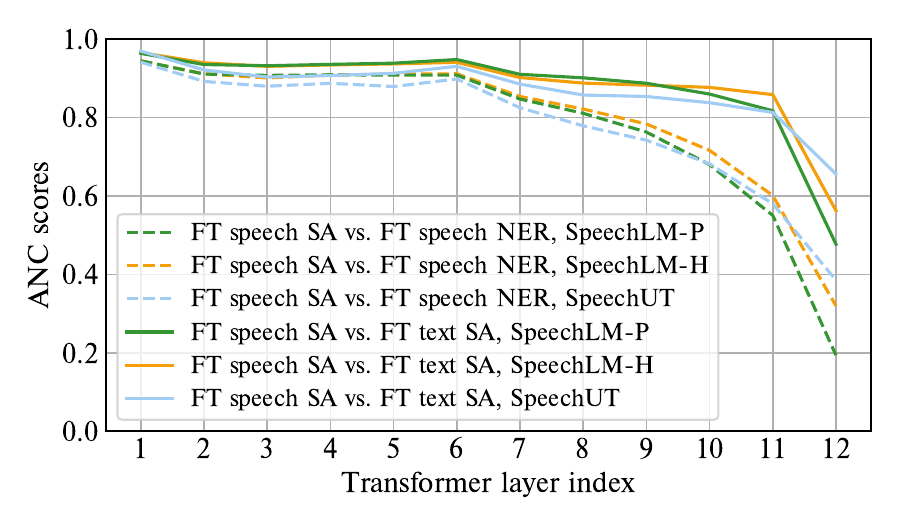}
    \vspace{-5mm}
    \caption{ANC scores between speech representations from models with different fine-tuning setups. \cmcremove{All of the curves are monotonically decreasing as we compare representations from different models and thus the error accumulates.} \kl{I don't understand what is meant by "the error accumulates".}}
    \label{fig:finetune_anc}
    \vspace{-1mm}
\end{figure}

In Fig.~\ref{fig:finetune_anc}, each curve corresponds to the ANC between speech representations from two models with different fine-tuning setups.
The solid lines compare models fine-tuned on the same task with different input \cmcedit{modalities} \cmcremove{forms (speech and text)}, while the dashed lines compare models fine-tuned on different tasks with the same input modality.
It can be observed that the solid lines are consistently higher than \kledit{the} dashed lines, which shows that the fine-tuning task affects the latent representations more than the input \cmcedit{modality}\cmcremove{form}.
This further supports the existence of knowledge transfer across different input \cmcedit{modalities}\cmcremove{forms}.

\begin{table*}[th!]
    \setlength{\tabcolsep}{4.5pt}
    \caption{$F_1$ scores (\%) of the Named Entity Recognition task on the SLUE dev set.}
    \footnotesize
    \centering
    \begin{tabular}{l c r r r r r r r r r r r r}
        \toprule
        \multicolumn{2}{c}{\multirow{2}{*}{Labeled Data
        }} & \multicolumn{6}{c}{without LM decoding} & \multicolumn{6}{c}{with LM decoding}\\
        \cmidrule(lr){3-8}
        \cmidrule(lr){9-14}
        & & \multicolumn{3}{c}{$F_1$ (\%)} & \multicolumn{3}{c}{Label-$F_1$ (\%)} & \multicolumn{3}{c}{$F_1$ (\%)} & \multicolumn{3}{c}{Label-$F_1$ (\%)}
        \\
        \cmidrule(lr){1-2}
        \cmidrule(lr){3-5}
        \cmidrule(lr){6-8}
        \cmidrule(lr){9-11}
        \cmidrule(lr){12-14}
        \multirow{2}{*}{Speech} & \multirow{2}{*}{Text} & Speech- & Speech- & Speech- & Speech- & Speech- & Speech- & Speech- & Speech- & Speech- & Speech- & Speech- & Speech- \\
        & & LM-P & LM-H & UT & LM-P & LM-H & UT & LM-P & LM-H & UT & LM-P & LM-H & UT \\
        \midrule
        \textit{Baselines} \\
        \hspace{2mm} 14.5 hrs & - & 64.2 & 64.0 & 62.9 & 76.4 & 78.2 & 77.2 & 73.0 & 73.1 & 72.8 & 81.8 & 82.1 & 81.8 \\ 
        \hspace{2mm} 14.5 hrs (syn) & - & 46.4 & 41.9 & 36.3 & 64.0 & 60.9 & 59.6 & 58.6 & 56.8 & 54.4 & 70.7 & 68.4 & 66.8 \\ 
        \midrule
        \textit{Proposed} \\
        \hspace{2mm} - & full & $^*$0.0 & $^*$1.2 & $^*$9.7 & $^*$0.2 & $^*$7.1 & $^*$32.8 & $^*$9.3 & $^*$29.4 & $^*$48.4 & $^*$9.4 & $^*$33.3 & $^*$58.3 \\
        \hspace{2mm} 10 mins & full & 35.6 & 34.7 & 31.5 & 45.6 & 45.8 & 46.5 & 50.4 & 51.7 & 51.8 & 56.5 & 58.1 & 59.3 \\
        \hspace{2mm} 1 hr & full & 50.2 & 52.0 & 47.3 & 65.1 & 65.9 & 64.9 & 62.5 & 63.4 & 63.3 & 71.7 & 72.2 & 72.4 \\
        \cmcedit{\hspace{2mm} 3 hrs} & full & 60.0 & 59.6 & 58.2 & 73.4 & 74.9 & 73.4 & 69.7 & 69.2 & 69.9 & 78.8 & 78.5 & 78.9 \\
        \bottomrule
        \multicolumn{10}{l}{\scriptsize$^*$ For text-only fine-tuning, we fine-tune the top 3 layers of the shared Transformer.}\\
        \multicolumn{10}{l}{\cmc{We need to reduce fontsize to make space for the fullnames of the models (i.e. SpeechLM \& SpeechUT).}}
    \end{tabular}
    \label{tab:NER_dev_results}
    \vspace{-3mm}
\end{table*}


\begin{table}[th!]
    \setlength{\tabcolsep}{1.2pt}
    \caption{$F_1$ scores (\%) of the Named Entity Recognition task on the SLUE test set with LM decoding.}
    \footnotesize
    \centering
    \begin{tabular}{l c r r r r r r}
        \toprule
        \multicolumn{2}{c}{Labeled Data
        } & \multicolumn{3}{c}{$F_1$ (\%)} & \multicolumn{3}{c}{Label-$F_1$ (\%)
        }\\
        \cmidrule(lr){1-2}
        \cmidrule(lr){3-5}
        \cmidrule(lr){6-8}
        \multirow{2}{*}{Speech} & \multirow{2}{*}{Text} & Speech- & Speech- & Speech- & Speech- & Speech- & Speech- \\
        & & LM-P & LM-H & UT & LM-P & LM-H & UT \\
        \midrule
        \textit{Baseline} \\
        \hspace{2mm} 14.5 hrs & - & \cmcedit{67.1}\cmcremove{73.0} & \cmcedit{67.0}\cmcremove{73.1} & \cmcedit{66.9}\cmcremove{72.8} & \cmcedit{75.8}\cmcremove{81.8} & \cmcedit{76.5}\cmcremove{82.1} & \cmcedit{75.2}\cmcremove{81.8} \\
        \midrule
        \textit{Proposed} \\
        \hspace{2mm} - & full & \cmcedit{$^*$8.1}\cmcremove{$^*$9.3} & \cmcedit{$^*$25.7}\cmcremove{$^*$29.4} & \cmcedit{$^*$45.1}\cmcremove{$^*$48.4} & \cmcedit{$^*$8.2}\cmcremove{$^*$9.4} & \cmcedit{$^*$30.0}\cmcremove{$^*$33.3} & \cmcedit{$^*$53.9}\cmcremove{$^*$58.3} \\
        \hspace{2mm} 1 hr & full & \cmcedit{56.9}\cmcremove{62.5} & \cmcedit{58.7}\cmcremove{63.4} & \cmcedit{58.3}\cmcremove{63.3} & \cmcedit{65.9}\cmcremove{71.7} & \cmcedit{67.7}\cmcremove{72.2} & \cmcedit{67.2}\cmcremove{72.4} \\
        \cmcedit{\hspace{2mm} 3 hrs} & full & \cmcedit{62.8}\cmcremove{69.7} & \cmcedit{63.8}\cmcremove{69.2} & \cmcedit{62.6}\cmcremove{69.9} & \cmcedit{72.8}\cmcremove{78.8} & \cmcedit{72.9}\cmcremove{78.5} & \cmcedit{72.0}\cmcremove{78.9} \\
        \midrule
        \midrule
        \multicolumn{8}{l}{\textit{Prior work (with 14.5 hrs labeled speech data)}} \\
        \hspace{2mm} W2V2~\cite{shon2022slue} & & & 63.4 & & & 71.7 & \\
        \hspace{2mm} HuBERT~\cite{shon2022slue} & & & 61.9 & & & 70.3 & \\
        \bottomrule
        \multicolumn{8}{l}{\scriptsize$^*$ For text-only fine-tuning, we fine-tune the top 3 layers of the shared Transformer.}\\
        \multicolumn{8}{l}{\cmc{We need to reduce fontsize to make space for the fullnames of the models (i.e. SpeechLM \& SpeechUT).}}\\
    \end{tabular}
    \label{tab:NER_test_results}
\end{table}

\begin{figure*}[t!]
    \centering
    \includegraphics[width=\linewidth]{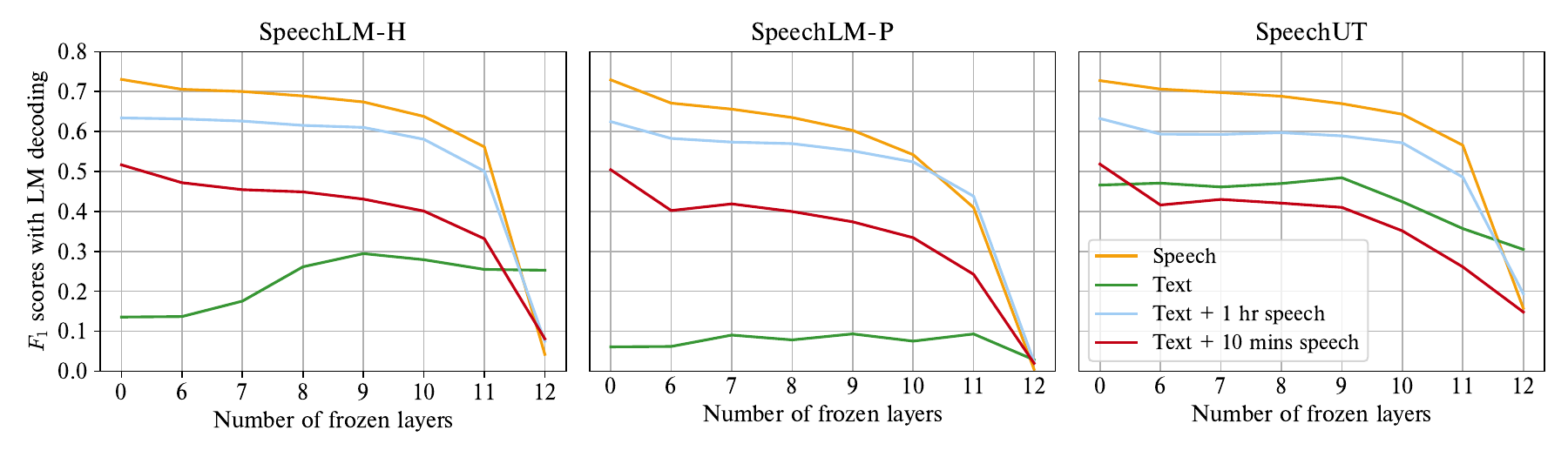}
    \vspace{-4mm}
    \caption{\(F_1\) scores \klremove{on}\kledit{for} NER on \kledit{the} SLUE dev \klremove{split}\kledit{set} with varying number of frozen layers during fine-tuning.
    }
    \label{fig:ner_frozen}
    \vspace{-1mm}
\end{figure*}

\vspace{-2.5mm}
\subsection{Fine-tuning with frozen bottom layers}
\vspace{-1.5mm}
\label{ssec:frozen}
In the previous experiments, we have evaluated the performance of speech-text models on few-shot and zero-shot SLU, as well as analyzed the latent representations of speech-text models and identified the ``first-align-then-predict" pattern.
We would then like to combine the two sets of observations to see whether we can further improve SLU performance by only fine-tuning the task-specific top layers.
We follow the setup in Sec. \ref{ssec:ner} to fine-tune speech-text models on NER, but with different numbers of bottom layers frozen.
The results are shown in Fig.~\ref{fig:ner_frozen}.
We find that by training only a few top layers (e.g., train 4 layers and leave 8 layers frozen), we can achieve a performance that is very close to the performance of 0 frozen layers, in both the few-shot setting and the full-speech fine-tuning setting.
This again aligns with the behavior of multi-lingual natural language models reported in the literature \cite{merchant2020what}.
On the other hand, in the zero-shot setting, the best performance is usually achieved with a certain number of bottom layers frozen (e.g., 9 layers for \cmcremove{SPLM}\cmcedit{SpeechLM}-H and \cmcremove{SPUT}\cmcedit{SpeechUT}). 
This supports our hypothesis that the bottom layers are in charge of representation alignment and thus should not be updated during fine-tuning for the best zero-shot transfer performance.


\section{Conclusions}
\vspace{-3mm}
In this work, we study the problem of zero-shot and few-shot spoken language understanding by fine-tuning speech-text models with labeled text data.
Our results demonstrate zero-shot transferability of pre-trained speech-text models from text to speech on these tasks.
We also show that, with only a small amount of labeled speech data, the performance can be significantly improved, almost matching previous work trained with a much larger amount of labeled speech on the SLUE benchmark.
Our analysis suggests that the bottom layers of speech-text models learn the alignment between speech and text representations, which is crucial to the model's performance in the absence of enough labeled speech data, while the top layers are task-specific and tend to be updated more during fine-tuning.
This analysis suggests freezing the bottom layers and only updating the top layers during fine-tuning, which results in the best performance under the zero-shot setting.

Our approach can be directly scaled up when more labeled speech/text data is available.
However, it is still an open question whether the model continues to benefit from text supervision when more \kledit{speech} data is available.
In addition, given the similarity we have observed between speech-text models and multi-lingual text models, it will also be interesting to study multilingual speech-text models with the methods introduced in this paper to see how the spoken and written forms of different languages can be integrated.

\kl{the references need another pass of proofreading.  I see some capitalization errors (USM, ASR, MSLAM, maybe others I missed).  Also double-check (if you haven't already) whether any of the preprints are now published somewhere.}
\vfill\clearpage
\pagebreak
\clearpage

\bibliographystyle{IEEEbib}
\bibliography{Template_Regular}

\end{document}